\documentclass[10pt,twocolumn,letterpaper]{article}

\usepackage{cvpr}              

\usepackage{graphicx}
\usepackage{amsmath}
\usepackage{amssymb}
\usepackage{booktabs}
\usepackage{bbding}
\usepackage{multirow}
\usepackage{multicol}
\usepackage{tablefootnote}
\usepackage{mathtools}
\usepackage{colortbl}
\definecolor{mygray}{gray}{.9}

\usepackage[pagebackref,breaklinks,colorlinks]{hyperref}

\usepackage[capitalize]{cleveref}
\crefname{section}{Sec.}{Secs.}
\Crefname{section}{Section}{Sections}
\Crefname{table}{Table}{Tables}
\crefname{table}{Tab.}{Tabs.}

\begin{document}

\title{MSMDFusion: Fusing LiDAR and Camera at Multiple Scales with \\ Multi-Depth Seeds for 3D Object Detection}
\author{
    Yang Jiao$^{1,2*}$ \quad  Zequn Jie$^{3}$\thanks{Equal contribution.} \quad Shaoxiang Chen$^{3}$ \quad Jingjing Chen$^{1,2}$ \quad Lin Ma$^{3}$ \quad Yu-Gang Jiang$^{1,2}$ \\
    $^1$ Shanghai Key Lab of Intell. Info. Processing, School of CS, Fudan University  \\ 
    $^2$ Shanghai Collaborative Innovation Center on Intelligent Visual Computing \\
    $^3$Meituan 
}

\maketitle

\begin{abstract}
Fusing LiDAR and camera information is essential for achieving accurate and reliable 3D object detection in autonomous driving systems. This is challenging due to the difficulty of combining multi-granularity geometric and semantic features from two drastically different modalities.
Recent approaches aim at exploring the semantic densities of camera features through lifting points in 2D camera images (referred to as ``\emph{seeds}'') into 3D space, and then incorporate 2D semantics via cross-modal interaction or fusion techniques.
However, depth information is under-investigated in these approaches when lifting points into 3D space, thus 2D semantics can not be reliably fused with 3D points. 
Moreover, their multi-modal fusion strategy, which is implemented as concatenation or attention, either can not effectively fuse 2D and 3D information or is unable to perform fine-grained interactions in the voxel space.
To this end, we propose a novel framework with better utilization of the depth information and fine-grained cross-modal interaction between LiDAR and camera, which consists of two important components.
First, a Multi-Depth Unprojection (MDU) method with depth-aware designs is used to enhance the depth quality of the lifted points at each interaction level. 
Second, a Gated Modality-Aware Convolution (GMA-Conv) block is applied to modulate voxels involved with the camera modality in a fine-grained manner and then aggregate multi-modal features into a unified space.
Together they provide the detection head with more comprehensive features from LiDAR and camera. 
On the nuScenes test benchmark, our proposed method, abbreviated as MSMDFusion, achieves state-of-the-art 3D object detection results with 71.5\% mAP and 74.0\% NDS, and strong tracking results with 74.0\% AMOTA without using test-time-augmentation and ensemble techniques. The code is available at https://github.com/SxJyJay/MSMDFusion.
\end{abstract}

\begin{figure}[!t]
\centering
\includegraphics[width=\columnwidth]{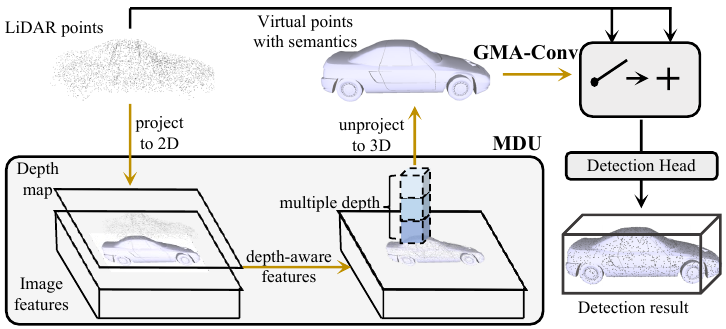} 
\caption{Illustration of our MSMDFusion pipeline. The yellow arrows indicate  information passing or interaction between the LiDAR and camera modalities.}
\label{fig1}
\end{figure}

\section{Introduction}
\label{sec:intro}

Detecting 3D objects~\cite{zhou2018voxelnet,yan2018second,bai2022transfusion} is regarded as a fundamental task for autonomous driving. Aiming at the robust environmental perception, LiDARs and cameras are widely equipped on autonomous driving vehicles since they can provide complementary information. Characterized by point clouds, LiDARs can capture accurate spatial information, while cameras contain rich semantics and context with images. 
Therefore, developing multi-modal detectors that enjoy the benefits of the two worlds is promising. Such an idea has catalyzed the emergence of a set of recent researches~\cite{chen2017multi,liang2018deep,qi2018frustum,yoo20203d,vora2020pointpainting,wang2021pointaugmenting,yin2021multimodal,bai2022transfusion,liu2022bevfusion,liang2022bevfusion}.
Early works~\cite{chen2017multi,liang2018deep,liang2019multi,sindagi2019mvx,yoo20203d,vora2020pointpainting,wang2021pointaugmenting,bai2022transfusion} perform LiDAR-camera fusion by projecting 3D LiDAR points (or region proposals generated from them) onto 2D image planes to collect useful 2D semantics. 
However, such a paradigm suffers from the signal density mismatching of multi-modality sensors. Since LiDAR points are much sparser than camera pixels, this projection manner will inevitably waste semantically rich 2D features~\cite{bai2022transfusion,liang2022bevfusion}.

Recently, another paradigm~\cite{li2022unifying,li2022voxel,yin2021multimodal,liu2022bevfusion,liang2022bevfusion} for LiDAR-camera fusion has emerged. Instead of collecting 2D semantic features with 3D queries, these methods first estimate depths of pixels, and then directly lift 2D pixels with their semantics to the 3D world (we refer to these pixels and corresponding lifted points as \textit{``seeds"} and \textit{``virtual points"} in this paper) to fuse with the real 3D point cloud. Two methods with the same name of BEVFusion~\cite{liu2022bevfusion,liang2022bevfusion} treat every image feature pixel as a seed and generate virtual points in the BEV space. MVP~\cite{yin2021multimodal} and VFF~\cite{li2022voxel} sample pixels from foreground regions and lift them to the voxel space. Benefited from the dense virtual points, such a paradigm not only maintains the semantic consistency in the image~\cite{liu2022bevfusion}, but also complements the geometric cues for the sparse LiDAR point cloud~\cite{yin2021multimodal}. 

Despite significant improvements have been made, existing methods in this line suffer from two major problems, which hampers benefiting from virtual points. 
First, depth, as the key to the quality of virtual points, is under-investigated in generating virtual points. On the one hand, depth directly determines the spatial location in 3D space of a seed via perspective projection which can thereby significantly affect 3D object detection results. 
On the other hand, depth can also enhance the semantics carried by virtual points by providing color-insensitive cues in describing objects~\cite{zhang2021depth}, 
since combining RGB information with depth guidance correlates camera pixels of similar depths and enables them to jointly contribute to capturing instance-related semantics when lifted as virtual points.
Existing multi-modal detectors~\cite{liu2022bevfusion,liang2022bevfusion,li2022unifying} mainly pay attention on interacting LiDAR points with camera virtual points, while ignoring the importance of seed depths in generating the virtual points.

Second, the fine-grained cross-modal interaction between virtual points and 3D points in the uncompressed space (e.g., the voxel space) is crucial but non-trivial. 
Generated virtual points are geometrically and semantically inconsistent with real LiDAR points due to imperfect depths and inherent modality gap.
Hence, in order to benefit from the semantically rich virtual points, it is necessary to adaptively select helpful information from virtual points under the guidance of real LiDAR points in a fine-grained and controllable manner. 
However, such a cross-modal interaction is constrained by the intensive memory and computation cost brought by the massive amounts and unstructured nature of point cloud data. 
Alternatively, existing approaches combine the multi-modal information with simple concatenate~\cite{yin2021multimodal} or add operations~\cite{li2022unifying} in the voxel space, or perform cross-attention in a compressed BEV space~\cite{yang2022deepinteraction}.

Aiming at unlocking the potential of virtual points and addressing the drawbacks of existing methods, we propose a multi-scale fusion framework, called MSMDFusion, and within each scale, there are two key novel designs, namely the Multi-Depth Unprojection (MDU) and Gated Modality-Aware Convolution (GMA-Conv). 
As shown in Fig.\ref{fig1}, MDU is mainly for enhancing the quality of generated virtual points in terms of geometric accuracy and semantic richness. 
To be specific, when lifting 2D seeds from image into the 3D space, multiple depths are explored within a reference neighborhood to generate virtual points with more reliable depth. Next, the camera feature and depth are combined to produce depth-aware features as stronger 2D semantics to decorate these virtual points. 
GMA-Conv takes real LiDAR points and generated virtual points as inputs, and performs fine-grained interaction in a \textit{select-then-aggregate} manner.
Concretely, we first adaptively select useful information from camera voxel features under the guidance of reference LiDAR voxels, then aggregate them grouped sparse convolutions for sufficient multi-modal interaction. We also specifically adopt a voxel subsampling strategy to efficiently obtain reliable LiDAR references when implementing our GMA-Conv.

Finally, with the resulting multi-modal voxel features from multiple scales, we further associate them with cascade connections across scales to aggregate multi-granularity information. 
With the above designs, the camera semantics encapsulated in the virtual points are consistently combined with LiDAR points, and thereby providing a stronger multi-modal feature representation for boosting the 3D object detection.
As shown in the Table~\ref{table:nvpf}, with 100 times fewer generated virtual points than the two BEVFusion methods~\cite{liang2022bevfusion,liu2022bevfusion}, our MSMDFusion can still achieve state-of-the-art performances. 

In summary, our contributions lie in threefold:
(1) We propose a novel MSMDFusion approach, which encourages sufficient LiDAR-Camera feature fusion in the multi-scale voxel space.
(2) Within each scale, we propose a Multi-Depth Unprojection strategy (MDU) to promote virtual points generation with better locations and semantics by fully leveraging depth of pixels, as well as a Gated Modality-Aware Convolution (GMA-Conv) to achieve fine-grained controllable multi-modal interaction.
(3) Extensive experimental results on the large-scale nuScenes dataset prove the effectiveness of our MSMDFusion and its components. We achieve state-of-the-art performances with 71.5\% mAP and 74.0\% NDS on the challenging nuScenes detection track using single model\footnote{We do not use Test-Time Augmentation (TTA) and model ensemble technique.}. When combining the simple greedy tracking strategy~\cite{yin2021center}, our method also achieves strong tracking results with 74.0\% AMOTA.

\begin{figure*}[!t]
\centering
\includegraphics[width=\linewidth]{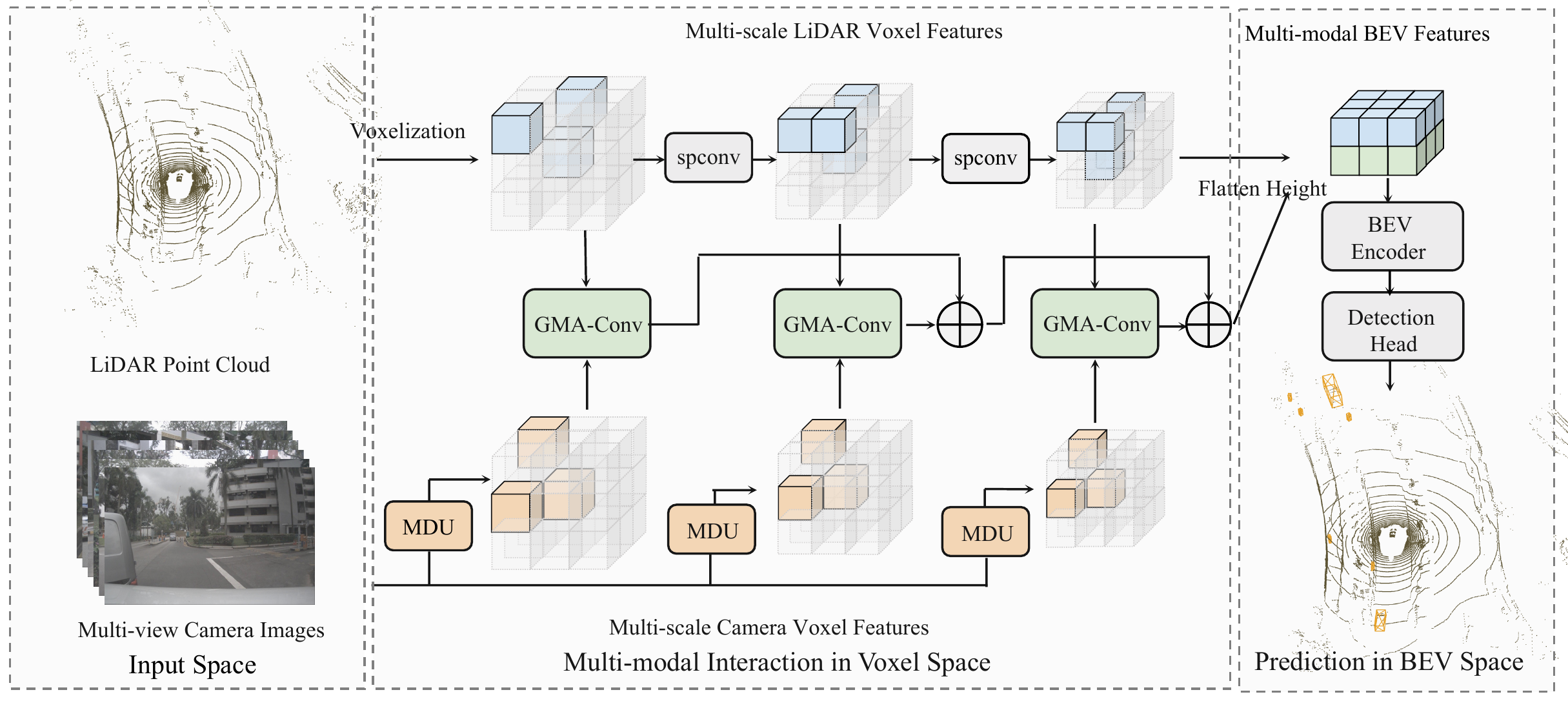}
\caption{The framework of our proposed MSMDFusion. 
Given a LiDAR point cloud and corresponding multi-view camera images, MSMDFusion first separately transforms them into the voxel space and obtains  multi-scale LiDAR and camera features. Then, the LiDAR and camera features at each scale interact through our proposed Multi-Depth Unprojection strategy (MDU) and  Gated Modality-Aware Convolution block (GMA-Conv, details in Fig.\ref{fig:MS-Conv}). The outputs of GMA-Conv at each scale are further aggregated for combining multi-granularity information. Finally, the resulting multi-modal and LiDAR voxel features are flattened into the BEV space for the final prediction. $\oplus$ represents voxel addition and ``spconv" indicates 3D sparse convolution.}
\label{fig:Framework}
\end{figure*}

\section{Related Work}
Recently, fusing LiDAR and camera signals in 3D detection has attracted increasing interest. Early works~\cite{chen2017multi,liang2018deep,vora2020pointpainting,yoo20203d,huang2020epnet,wang2021pointaugmenting} project 3D queries to 2D camera images for collecting useful semantics. Under such a paradigm, MV3D~\cite{chen2017multi} and AVOD~\cite{ku2018joint} associate 3D proposals with 2D RoI features. While PointPainting~\cite{vora2020pointpainting} and PointAugmenting~\cite{wang2021pointaugmenting} directly decorate raw 3D points with 2D semantics. EP-Net~\cite{huang2020epnet} and 3D-CVF~\cite{yoo20203d} perform multi-modal fusion at both point level and proposal level. However, since 3D points are inherently sparse, such a hard association approach wastes the dense semantic information in 2D features. 
Recently, multi-modal 3D detectors~\cite{yin2021multimodal,liang2022bevfusion,liu2022bevfusion,yang2022deepinteraction} lift dense 2D seeds to 3D space for learning the 2D-3D joint representation in a shared space. Two BEVFusion methods~\cite{liang2022bevfusion,liu2022bevfusion} densely lift every image feature pixel into 3D space and then encode these lifted points as another BEV (bird’s-eye view) map to fuse with the BEV map obtained using the original 3D point cloud. LiDAR and camera signals will finally share a common representation space, but they do not interact at any stage before BEV fusion. MVP~\cite{yin2021multimodal} selects seeds from foreground objects and unprojected them into the LiDAR frame for points cloud augmentation. With the seeds sampling strategy similar to MVP, UVTR~\cite{li2022unifying} constructs both LiDAR and camera branches in the voxel space to maintain the modality-specific information, and then interact them with voxel add operation. However, the LiDAR and camera branches are constructed in a single scale of voxel space, losing the benefits of different levels of feature abstraction in multi-scale voxel space. Moreover, merging LiDAR and camera voxels with simple addition limits their fine-grained interaction.

\section{Method}
\subsection{Framework Overview}
An overall view of MSMDFusion is shown in Fig.\ref{fig:Framework}. Given a LiDAR point cloud and corresponding multi-view camera images as inputs, MSMDFusion first extracts multi-scale features from both modalities in the voxel space. Then, LiDAR-camera interaction is performed within multi-scale voxel space to properly combine multi-granularity information from both modalities.
At each scale, we specifically design a Multi-Depth Unprojection (MDU) strategy to obtain high-quality virtual points in the voxel space, and a  Gated Modality-Aware Convolution block (GMA-Conv) for effective LiDAR-camera interaction and fusion.
We also introduce cross-scale connections to progressively combine features of different granularities. 
Afterward, the deeply interacted multi-modal features, together with the LiDAR features, are transformed into the BEV space and fed to the BEV encoder and detection head for the final prediction.

\subsection{LiDAR and Camera Feature Extraction}
We first extract high-level features from the raw inputs of LiDAR and camera.
For a given LiDAR point cloud, we voxelize the points and extract their features in the voxel space using a set of 3D sparse convolutional blocks~\cite{graham20183d} following prior arts~\cite{bai2022transfusion,liu2022bevfusion,liang2022bevfusion}. Each convolutional block outputs voxel features of different scales, representing different levels of abstraction for point cloud, and these multi-scale features will further interact with their 2D counterparts through our proposed GMA-Conv blocks.

For the multi-view camera images, we adopt ResNet50 with FPN as the image backbone to extract multi-scale image features, which contains rich object semantics. However, it is non-trivial to transfer these useful semantics to the 3D voxel field, since images are inherently 2D data. 
Motivated by recent works~\cite{yin2021multimodal,liu2022bevfusion,liang2022bevfusion} which select seeds from the image plane and estimate their depths to lift them as 3D virtual points, we further propose a Multi-Depth Unprojection strategy to lift 2D seeds, which could mitigate the drawbacks of prior works.
The details will be presented in the following subsections.

\begin{figure}[!t]
\centering
\includegraphics[width=\columnwidth]{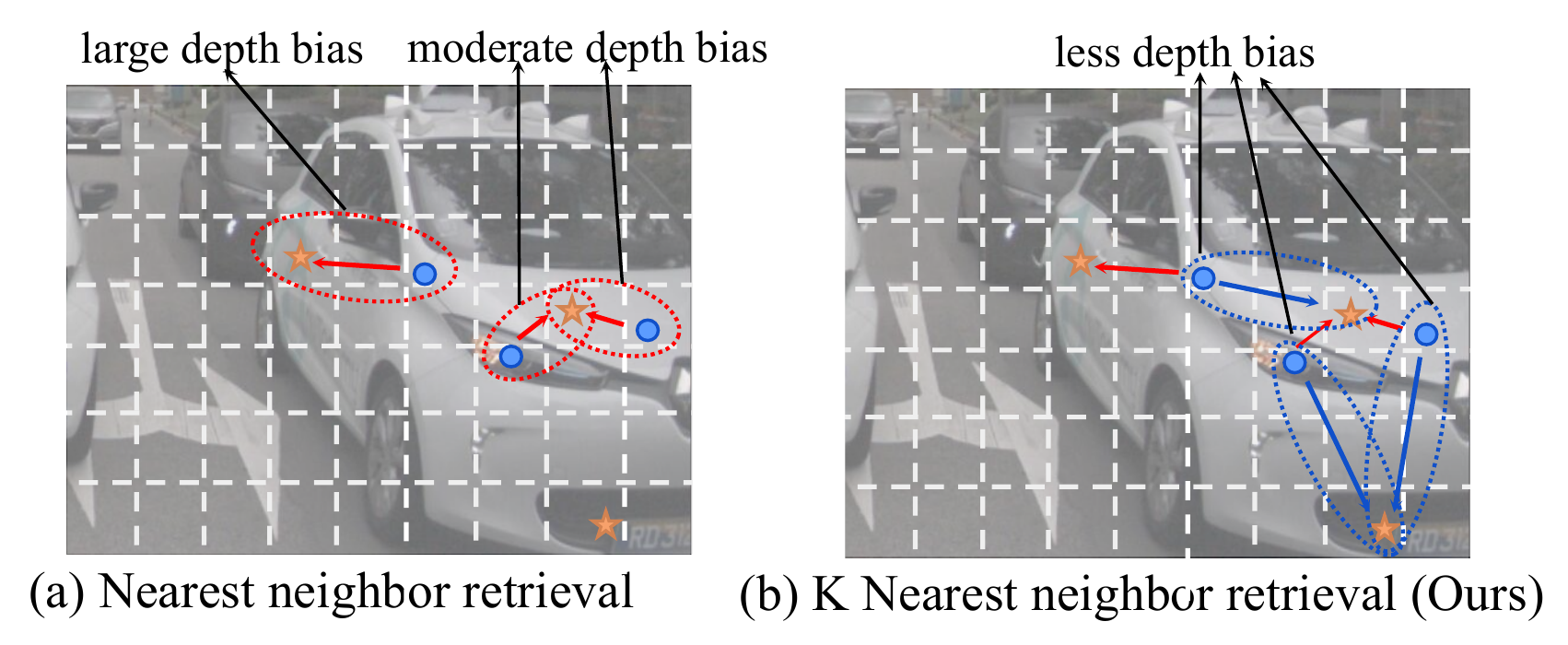}
\caption{Comparison of the effects of retrieving reference points from nearest neighbor (MVP) and K-nearest neighbor (Ours). The blue circle and yellow star represent seeds and reference points, respectively. To keep the figure concise, we only depict the case that K=2. Our strategy can obtain more reliable depths by exploring more neighbors.}
\label{fig:NNvsKNN}
\end{figure}

\subsection{Multi-Modal Interaction in Voxel Space}
The extracted LiDAR and camera features from the previous stage respectively focus on geometric and semantic information, hence the goal of multi-modal interaction is to properly fuse these features into a unified rich representation of the scene. As shown in the Fig.\ref{fig:Framework}, the components of LiDAR-camera interaction within each scale in the voxel space are conceptually simple: 1) Multi-Depth Unprojection (MDU) strategy is introduced to improve the quality of the generated virtual points and 2) Gated Modality-Aware Convolution (GMA-Conv) block is designed for controllable LiDAR-camera fusion. 
In the following parts, we will first elaborate on MDU and GMA-Conv within a specific scale, then on top of them, we describe the complete multi-scale fusion process.

\subsubsection{Multi-Depth Unprojection}
\label{mdu}
To lift a 2D seed in the pixel space to the 3D space (i.e., the unprojection operation), the depth associated with each seed should be estimated~\cite{wang2019pseudo}. To obtain an initial reliable depth estimation, we use the strategy from MVP~\cite{yin2021multimodal} as the basis as shown in Fig.\ref{fig:NNvsKNN}(a).
Formally, given a point cloud $\mathcal{P}=\{(x_i,y_i,z_i)\}_{i=1}^{N_P}$ and multi-view images as inputs, MVP first projects 3D points onto 2D images and preserves points falling within the 2D foreground instance masks. These 3D points are kept as reference points to provide depth, and they are denoted as $\mathcal{R}=\{(u_i,v_i,d_i)\}_{i=1}^{N_R}$, where $u_i$ and $v_i$ are pixel coordinates and $d_i$ is the real depth. Then, a set of seeds $\mathcal{S}=\{(u^s_i,v^s_i)\}_{i=1}^{N_S}$ are uniformly sampled from each instance mask on the image, and each of them will retrieve a real depth from its nearest reference point as its estimated depth. Finally the sampled seeds with estimated depths are unprojected to 3D as virtual points.

Though effective, such a strategy ignores the fact that spatial proximity in 2D images can not be guaranteed in 3D, 
and this may lead to inaccurate depth estimation as shown in the red circle of Fig.\ref{fig:NNvsKNN}(a).
Toward this end, we propose to equip every seed with multiple depths by retrieving K-nearest reference points, which can be regarded as a soft strategy to achieve more reliable depth estimation. 
As shown in the green circle of the Fig.\ref{fig:NNvsKNN}(b), each seed can be unprojected using multiple (K) depths from neighboring reference points, which generates K virtual points to improve its recall of the actual 3D points. We provide more detailed discussion on the effect of multi-depth seeds in Sec.\ref{mdu_discussion}.

\begin{figure*}[!t]
\centering
\includegraphics[width=0.8\linewidth]{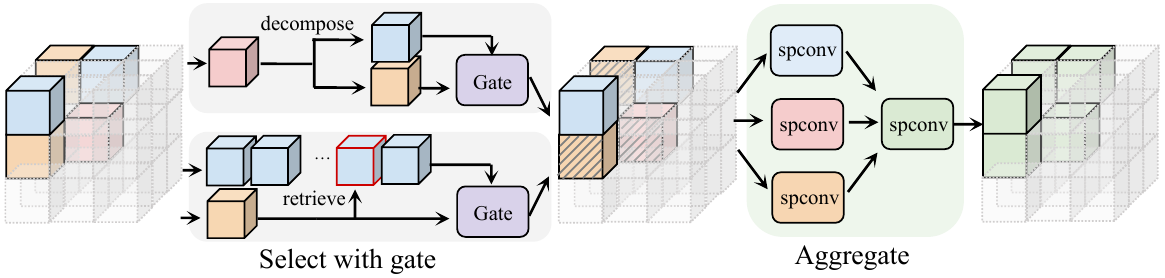}
\caption{Illustration of the select-then-aggregate process in the Gated Modality-Aware Convolution. Colors indicate different modalities: yellow for camera, blue for LiDAR, and red for LiDAR-camera combined. Cubes with shadows represent voxels selected with gates.}
\label{fig:MS-Conv}
\end{figure*}

These generated virtual points will be further decorated to make the best use of semantics encapsulated in images.
But different from MVP~\cite{yin2021multimodal} that simply decorates virtual points with class scores predicted by a pretrained 2D detector, we make an early interaction between image features and sparse depth information from reference points to generate depth-aware semantic features.
These depth-aware semantic features are then used to decorate virtual points in an adaptive manner controlled by their corresponding seed depths. The whole process can be trained end-to-end.
Concretely, we concatenate the camera image feature $C$ with a sparse depth map containing all depths of reference points $\mathcal{R}$ in that image, and then fuse them with a convolutional layer to obtain the depth-aware semantic feature $C^{d}$. 
For the K virtual points generated by the same seed, the semantic feature of the corresponding pixel should also contribute differently along the projection ray in the 3D space~\cite{philion2020lift}.
Thus we compute a dynamic weight factor for each individual depth to modulate the semantic feature.
The process for a specific seed $(u^{s}_{i},v^{s}_{i})$ can be formally described as:
\begin{equation}
\begin{split}
    & s^{k}_{i} = \mathtt{Sigmoid}(\mathtt{Linear}([C^{d}[u^s_i, v^s_i]; d^{s,k}_{i}]))\\
    & c^{k}_{i} = C^{d}[u^s_i, v^s_i] * s^{k}_{i} \\
\end{split}
\end{equation}
where $d^{s,k}_{i}$ is the $k$-th estimated depth of seed $(u^{s}_{i},v^{s}_{i})$, $s^{k}_{i}\in [0,1]$ represents its weight factor, and $c^{k}$ is the modulated semantic feature, which will be used to decorate the corresponding unprojected virtual point $(u^{s}_{i},v^{s}_{i},d^{s,k}_{i})$. 

The resulting virtual points with decorated features will be voxelized to match the output resolution of the corresponding scale of LiDAR branch, in order to perform interaction in the voxel space.

\subsubsection{Gated Modality-Aware Convolution} \label{sec:GMA-Conv}
With the resulting LiDAR and camera voxel features that share the same spatial resolution (i.e., the vertically aligned blue and yellow voxels in Fig.\ref{fig:Framework}), we aim to interact them in a fine-grained and controllable manner with our proposed Gated Modality-Aware Convolution (GMA-Conv) block in each scale. 
As shown in Fig.\ref{fig:MS-Conv}, we first group the voxels according to their modalities: LiDAR-only (blue voxels), camera-only (yellow voxels), as well as LiDAR and camera combined (red voxels), and they are denoted as $f^L$, $f^C$, and $f^{LC}$, respectively. 
Then, more fine-grained multi-modal interaction is conducted in a \textit{select-then-aggregate} manner. 

\textbf{Select.} Motivated by the fact that LiDAR-based detectors usually surpass their camera-based counterparts with a large margin, we take LiDAR as the guiding modality to select useful information from camera features. Specifically, we design a gate conditioned on the LiDAR feature to control and update its camera counterpart, which can be formally described as:
\begin{equation}
\label{eq:gate}
    \tilde{f^{C}_{i}} = \mathtt{ReLU}(\mathtt{Linear}(f^{L}_{j})) * f^{C}_{i}
\end{equation}
where $\tilde{f^{C}_{i}}$ is the updated camera voxel feature, and $i$ and $j$ are voxel indices of paired camera and LiDAR voxel features ($f^{L}_{j}$ can be regarded as the reference voxel for $f^{C}_{i}$). 
The key to make this selection process effective is how to efficiently find reliable reference voxels.
A straightforward way is to retrieve the nearest LiDAR voxels for each camera-only voxel $f^{C}_{i}$ from $f^{L}$. And similarly for LiDAR-camera voxels $f^{LC}_{k}$, the reference can be retrieved from both the LiDAR and camera modalities (i.e., $f^{L}$ and $f^{C}$).
Although nearby neighbors from more reliable modalities can serve as a good reference, such a retrieval process is infeasible to be implemented with exhaustive pairwise distance calculation, because the computation and memory cost brought by massive amounts of LiDAR and camera voxels is intractable, and we will give more analysis and our solution at the end of this subsection. 

\textbf{Aggregate.} With the original LiDAR-only voxels $f^{L}$, as well as the updated camera-only and LiDAR-camera combined voxels $\tilde{f}^{C}$ and $\tilde{f}^{LC}$, we first transform their modality-specific representation into an intermediate joint space with 3D sparse convolutions. 
Then, within the joint space, we further combine all voxels and promote their interaction through another 3D sparse convolution as shown in Fig.\ref{fig:MS-Conv}. The resulting multi-modal voxel features are denoted as $F^M$. 

\begin{figure}[!t]
\centering
\includegraphics[width=\columnwidth]{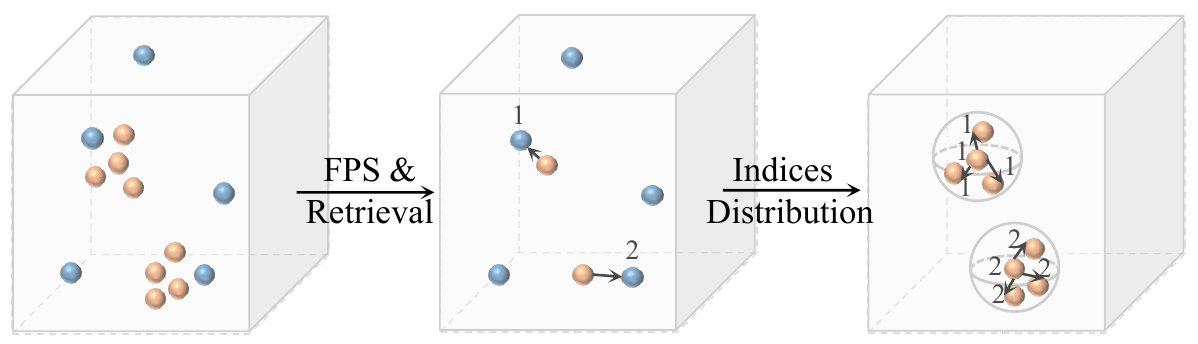}
\caption{The pipeline of our implementation of the retrieval process. For concise illustration, voxel features are represented by balls of different colors: yellow for camera and blue for LiDAR.}
\label{fig:Impl}
\vspace{-7mm}
\end{figure}

\textbf{Efficient Implementation.} 
We first discuss two operators commonly used for cross-modal interaction to demonstrate their complexity bottleneck. Then we present our implementation of the retrieval process aforementioned. 
Suppose that the numbers of LiDAR and camera voxels are N and M, respectively, where N and M are generally at the scale of 10$^5$ for a LiDAR frame. Hence, any operation with quadratic computation or space complexity (i.e., $\mathcal{O}(\rm{N}^2)$, $\mathcal{O}(\rm{M}^2)$, or $\mathcal{O}(\rm{MN})$) is intractable, e.g., the standard cross-attention~\cite{vaswani2017attention}.
Alternatively, local attention introduced in Swin-Transformer~\cite{liu2021swin} seems to be a feasible solution. 
However, unlike 2D data, voxels are not stored as dense tensors where nearby elements can be easily accessed by index. Instead, pairwise distances have to be calculated to retrieve local neighbors for each voxel. This process again leads to quadratic complexity.

Our efficient implementation of the aforementioned process is shown in Fig.\ref{fig:Impl}. The acceleration is achieved by taking advantage of the fact that camera voxels are generally redundant due to the dense unprojection as described in Sec.\ref{mdu}, we first subsample L elements from $f^{C}$ with the Farthest Point Sampling (FPS)~\cite{qi2017pointnet++} strategy. Then, we retrieve nearest LiDAR voxels for these L camera voxels with exhaustive pairwise distance calculation. 
Finally, each of these L camera voxels distribute its nearest LiDAR voxel's index to neighbors within the ball of a predefined radius, i.e., these camera voxels share a common LiDAR voxel as reference in the selection process.
The overall computation complexity of our implementation is $\mathcal{O}(\rm LM+LN+LN)=\mathcal{O}(L(\rm M+N))$,
since L is a small constant compared with N and M (we set L to 2048 in practice), and the 3D sparse convolution is efficiently implemented with complexity linear to N and M, thus the overall computation complexity of our GMA-Conv is $\mathcal{O}(\rm M+N)$.

\begin{table*}[!t]
\centering
\scalebox{0.9}{
\begin{tabular}{l|c|cc|cccccccccc}
\toprule
\multicolumn{1}{l|}{Method} & Modality & mAP  & NDS   & Car  & Truck & C.V. & Bus  & Trailer & Barrier & Motor. & Bike & Ped. & T.C. \\ \midrule
PointPillar~\cite{lang2019pointpillars}                 & L        & 40.1 & 55.0 & 76.0 & 31.0  & 11.3 & 32.1 & 36.6    & 56.4    & 34.2   & 14.0 & 64.0 & 45.6 \\
CenterPoint~\cite{yin2021center}                 & L        & 60.3 & 67.3 & 85.2 & 53.5  & 20.0 & 63.6 & 56.0    & 71.1    & 59.5   & 30.7 & 84.6 & 78.4 \\
TransFusion-L~\cite{bai2022transfusion}               & L       & 65.5 & 70.2 & 86.2 & 56.7  & 28.2 & 66.3 & 58.8    & 78.2    & 68.3   & 44.2 & 86.1 & 82.0 \\
PointPainting~\cite{vora2020pointpainting}               & LC       & 46.4 & 58.1 & 77.9 & 35.8  & 15.8 & 36.2 & 37.3    & 60.2    & 41.5   & 24.1 & 73.3 & 62.4 \\
3D-CVF~\cite{yoo20203d}                      & LC     & 52.7 & 62.3 & 83.0 & 45.0  & 15.9 & 48.8 & 49.6    & 65.9    & 51.2   & 30.4 & 74.2 & 62.9 \\
PointAugmenting~\cite{wang2021pointaugmenting}             & LC       & 66.8 & 71.0 & 87.5 & 57.3  & 28.0 & 65.2 & 60.7    & 72.6    & 74.3   & 50.9 & 87.9 & 83.6 \\
MVP~\cite{yin2021multimodal}                         & LC     & 66.4 & 70.5 & 86.8 & 58.5  & 26.1 & 67.4 & 57.3    & 74.8    & 70.0   & 49.3 & 89.1 & 85.0 \\
TransFusion~\cite{bai2022transfusion}                 & LC     & 68.9 & 71.7 & 87.1 & 60.0  & 33.1 & 68.3 & 60.8    & 78.1    & 73.6   & 52.9 & 88.4 & 86.7 \\
VFF~\cite{li2022voxel}                 & LC     & 68.4 & 72.4 & 86.8 & 58.1  & 32.1 & 70.2 & 61.0    & 73.9    & \textbf{78.5}   & 52.9 & 87.1 & 83.8 \\
BEVFusion~\cite{liang2022bevfusion}                   & LC    & 69.2 & 71.8 & 88.1 & 60.9  & 34.4 & 69.3 & 62.1    & 78.2    & 72.2   & 52.2 & 89.2 & 85.2 \\
BEVFusion~\cite{liu2022bevfusion}                   & LC      & 70.2 & 72.9 & \textbf{88.6} & 60.1  & \textbf{39.3} & 69.8 & 63.8    & 80.0    & 74.1   & 51.0 & 89.2 & 86.5 \\
UVTR~\cite{li2022unifying}                 & LC     & 67.1 & 71.1 & 87.5 & 56.0  & 33.8 & 67.5 & 59.5    & 73.0    & 73.4   & 54.8 & 86.3 & 79.6 \\
DeepInteraction~\cite{yang2022deepinteraction}                 & LC     & 70.8 & 73.4 & 87.9 & 60.2  & 37.5 & 70.8 & 63.8    & 80.4    & 75.4   & 54.5 & 90.3 & 87.0 \\ \midrule
MSMDFusion (Ours)                        & LC     & \textbf{71.5}    & \textbf{74.0}    & 88.4    & \textbf{61.0}     & 35.2    & \textbf{71.4}    & \textbf{64.2}       & \textbf{80.7}       & 76.9    & \textbf{58.3}    & \textbf{90.6}   & \textbf{88.1}    \\ \bottomrule
\end{tabular}}
\caption{Comparison with state-of-the-art methods on nuScenes test set. Note that these are all single-model results without ensemble or test-time augmentation. We highlight the best performances across all methods with \textbf{bold}.}
\label{table:sota}
\end{table*}

\begin{table*}[!h]
\centering
\scalebox{0.9}{
\begin{tabular}{l|c|cccccc}
\toprule
Method      & Modality & AMOTA $\uparrow$ & AMOTP $\downarrow$ & TP $\uparrow$ & FP $\downarrow$ & FN $\downarrow$ & Recall $\uparrow$\\ \midrule
CenterPoint~\cite{yin2021center} & L        & 63.8  & 0.606 & 95877 & 18612 & 22928 & 67.5\\
TransFusion~\cite{bai2022transfusion} & LC       & 71.8  & 0.551 & 96775 & 16232 & 21846 & 75.8 \\
UVTR~\cite{li2022unifying}   & LC       & 70.1     & 0.686  & 98434 & 15615 & 20190 &75.0   \\
\midrule
\rowcolor{mygray} BEVFusion~\cite{liu2022bevfusion}* & LC & 74.1 & 0.403 & 99664 & 19997 & 19395 & 77.9 \\ \midrule
MSMDFusion (Ours)  & LC       & \textbf{74.0}  & \textbf{0.549} & \textbf{98624} & \textbf{14789} & \textbf{19853} & \textbf{76.3} \\ \bottomrule
\end{tabular}}
\caption{Comparison with leading detectors that use a greedy tracker on nuScenes tracking task. * represents that BEVFusion uses an ensemble model, while we only use a single model. We use \textbf{bold} to denote the best performances among all approaches with a single model.}
\label{table:tracking}
\vspace{-3mm}
\end{table*}

\subsubsection{Multi-Scale Progressive Interaction}
After performing MDU and GMA-Conv within different scales in the voxel space, multi-modal voxel features $\{F^M_i\}$ are generated for each scale. To further aggregate these multi-granularity information from $\{F^M_i\}$, we introduce cascade connections across different scales as shown in Fig.\ref{fig:Framework}, which can be formulated as:
\begin{equation}
\label{eq4}
    \hat{F}^M_{i+1} = F^M_{i+1} + \mathtt{DownSample}(\hat{F}^M_i),
\end{equation}
where the voxel downsample operation $\mathtt{DownSample}(\cdot)$ is applied to align their spatial resolutions, and $\hat{F}^M_{i+1}$ is the resulting voxel features that combine multi-scale information from the current and previous scales.

With the above designs, multi-granularity LiDAR-camera features can be thoroughly interacted at multiple scales in the voxel space.
The final output multi-modal voxel features can serve as a powerful representation for the subsequent modules.

\subsection{Prediction in BEV Space}
Following the workflow of prevalent voxel-based 3D object detectors~\cite{yan2018second,zhou2018voxelnet,yin2021center}, we compress the height of voxel features to transform them into the BEV (bird’s-eye view) space.
Specifically, we first fuse the LiDAR and multi-modal BEV features with a lightweight 2D convolution block. Then, the BEV feature containing enhanced multi-modality information is fed into a conventional BEV encoder and a detection head for the final prediction.

\section{Experiments}
\subsection{Experimental Setup}
\subsubsection{Dataset and Metrics}
\vspace{-1mm}

The nuScenes~\cite{nuscenes} dataset is a large-scale autonomous driving benchmark including 10,000 driving scenarios in total, which are split into 700, 150, and 150 scenes for training, validation, and testing, respectively. For detection, nuScenes defines a set of evaluation protocols, including the nuScenes Detection Score (NDS), mean Average Precision (mAP), as well as five True Positive (TP) metrics, namely mean Average Translation Error (mATE), mean Average Scale Error (mASE), mean Average Orientation Error (mAOE), mean Average Velocity Error (mAVE) and mean Average Attribute Error (mAAE). We report mAP, which is the mean of the average precision across ten classes under distance thresholds of 0.5m, 1m, 2m, 4m. NDS is the weighted combination of mAP, mATE, mASE, mAOE, mAVE and mAAE.

\noindent \textbf{Implementation Details} We use ResNet-50~\cite{he2016deep} with FPN~\cite{lin2017feature} as the image backbone and VoxelNet~\cite{yan2018second} as the LiDAR backbone. We set the image size to 448$\times$800, and voxel size as (0.075\textit{m}, 0.075\textit{m}, 0.2\textit{m}) following~\cite{bai2022transfusion}. To make the best use of multi-scale semantics, we extract features from four levels of the FPN (C2 to C5). Following MVP~\cite{yin2021multimodal}, we use CenterNet2~\cite{zhou2021probabilistic} as 2D detector when generating virtual points, and 50 seeds are selected on each instance unless otherwise specified. Our model training has two stages: (1) We first train a LiDAR-only detector for 20 epochs as our 3D backbone. (2) We then connect the proposed LiDAR-camera fusion modules with the 3D backbone for a joint training of another 6 epochs. 
The data augmentation strategies and training schedules are the same as prior works~\cite{yin2021center,zhu2019class}. We do not use Test-Time Augmentation (TTA) or multi-model ensemble during inference.

\begin{table}[!t]
\centering
\scalebox{0.9}{
\begin{tabular}{l|c|cc|c}
\toprule
Method & NVPF\tablefootnote{NVPF of BEVFusion~\cite{liang2022bevfusion} is calculated with its official code. BEVFusion~\cite{liu2022bevfusion} reports its NVPF in the paper. The former has a higher NVPF because it generates virtual points from image features of larger resolution.} & mAP & NDS & FPS\\ \midrule
BEVFusion~\cite{liang2022bevfusion} & 5M & 69.2   & 71.8  & 0.7 \\
BEVFusion~\cite{liu2022bevfusion}   & 2M & 70.2   &72.9  & \textbf{8.4} \\ \midrule
MSMDFusion(Ours) & 16k  & \textbf{71.5}   & \textbf{74.0} & 2.1  \\ \bottomrule
\end{tabular}}
\caption{Number of Virtual points Per Frame (NVPF) and performance comparison with two strongest methods on nuScenes test set. The FPS is tested on a single NVIDIA RTX 3090 GPU.}
\label{table:nvpf}
\end{table}

\subsection{Comparison with State-of-the-art}
We compare our MSMDFusion with state-of-the-art approaches on the nuScenes test set. Overall, Table~\ref{table:sota} shows that our method surpasses all existing methods, and achieves the new state-of-the-art performances of 71.5 mAP and 74.0 NDS. Meanwhile, our method maintains consistent performance advantages on most object categories, especially on the challenging category: \textit{Bike}, where a 3.5\% absolute AP gain over the strongest competitor is achieved.

Since our MSMDFusion and two strong BEVFusion methods all fuse LiDAR and camera signals via generating 3D virtual points from 2D seeds, we specifically compare our method with them in terms of the number of generated virtual points per LiDAR frame as shown in Table~\ref{table:nvpf}. The results suggest that although with 100 times fewer generated virtual points (16k vs 2M/5M) than them, our MSMDFusion still outperforms them in terms of both mAP and NDS, which proves that our method can better utilize 2D semantics. Meanwhile, we also compare the FPS of our method and two BEVFusion approaches. BEVFusion~\cite{liu2022bevfusion} can achieve the fastest inference speed due to its efficient camera-to-BEV transformation implemented with CUDA. Without such a hardware-friendly implementation, BEVFusion~\cite{liang2022bevfusion} suffers from processing a huge amount of virtual points, while our MSMDFusion runs much faster although performing multi-scale fine-grained fusion in the voxel space.

Besides, to demonstrate that our MSMDFusion can also generalize to other downstream tasks, we further perform the tracking task on nuScenes. For fair comparison, we only compare with single-model methods (i.e., without TTA or ensemble) using the greedy tracker introduced in~\cite{yin2021center}. As shown in Table~\ref{table:tracking}, our method consistent outperforms other single model competitors across all evaluation metrics. Meanwhile, although without using ensemble techniques, our method also achieve competitive results with the ensemble BEVFusion~\cite{liu2022bevfusion} method.

\subsection{Comprehensive Analysis}
Following UVTR~\cite{li2022unifying}, we conduct the following ablation studies on a randomly sampled 1/4 split of nuScenes training set for efficiency.
\subsubsection{Ablation of proposed components}
We conduct comprehensive ablation studies for each of our proposed components as shown in  Table~\ref{table:abl}. We use TransFusion-L~\cite{bai2022transfusion} as our baseline (\#1), and perform multi-scale fusion across all experiments. 

From the results of the Table~\ref{table:abl}, we have the following observations. 
(i) Introducing virtual points for multi-modal interaction (\#2 and \#3) brings evident improvements over the baseline, which proves that fusing dense 2D semantics from virtual points with LiDAR features in the voxel space can significantly enhance the LiDAR-only detector.
(ii) An early depth-color interaction (\#2 and \#3) can provide stronger semantics, which further boosts the benefits of interaction between LiDAR and virtual points.
(iii) Using GMA-Conv to select helpful information from virtual points (\#4 and \#5) can bring further improvements over absorbing all information from virtual points without distinction (\#2 and \#3).
(iv) With all these components combined (\#5), the final mAP and NDS are significantly improved from 60.26\% to 66.93\% and 65.62\% to 68.93\%, respectively, which proves the effectiveness of our designs.

\begin{table}[!t]
\centering
\scalebox{0.9}{
\begin{tabular}{c|ccc|cc}
\toprule
\# & MDU* & MDU & GMA-Conv & mAP & NDS \\ \midrule
1  &      &     &          & 60.26   & 65.62    \\
2  & \checkmark    &     &          & 64.33    & 67.03     \\
3  &      & \checkmark   &          & 65.59    & 68.04     \\
4  & \checkmark    &     & \checkmark        & 65.97    & 68.24    \\
5  &      & \checkmark   & \checkmark        & \textbf{66.93}    & \textbf{68.93}    \\ \bottomrule
\end{tabular}}
\caption{Ablation studies for our proposed components on the nuScenes validation set. MDU* is a degraded version of MDU by discarding the depth-aware feature generating process.}
\label{table:abl}
\end{table}

\begin{table}[!t]
\centering
\scalebox{0.9}{
\begin{tabular}{c|cccc|cc}
\toprule
\# &C2 & C3 & C4 & C5 & mAP & NDS \\ \midrule
1&   & & & & 60.26 & 65.62 \\ 
2&\checkmark  &    &    &    & 65.51   & 67.92   \\
3&   & \checkmark  &    &    & 64.99   & 67.97   \\
4&   &    & \checkmark  &    & 64.97   & 67.66   \\
5&   &    &    & \checkmark  & 64.52  & 67.61   \\
6&\checkmark  & \checkmark  &    &    & 66.34   & 68.63   \\
7&\checkmark  & \checkmark  & \checkmark  &    & 66.80   & 68.86   \\
8&\checkmark  & \checkmark  & \checkmark  & \checkmark  & \textbf{66.93}   & \textbf{68.93}   \\ \bottomrule
\end{tabular}}
\caption{Effects of LiDAR-camera interaction at different scales, C2-C5 represent the scales where the spatial resolutions of the features decrease from C2 to C5.}
\label{table:scale}
\vspace{-3mm}
\end{table}

\subsubsection{Effects of multi-scale interaction}
We also investigate the effects of performing LiDAR-camera interaction at different scales. 
As shown in Table~\ref{table:scale}, for models with single-scale interaction (\#2-\#4), interacting at scales of larger spatial resolution generally leads to larger performance gains over the baseline (\#1), which mainly benefits from more fine-grained image and LiDAR features. 
If the multi-modal interaction is performed at more scales (\#6-\#8), multi-granularity information can be integrated into our progressive interaction process to consistently achieve better performances than single-scale models. The complete multi-scale interaction model achieves performance gains of $\sim$2\% mAP and $\sim$1\% NDS over the best single-scale model, which indicates that multi-scale interaction is beneficial for comprehensive multi-modal fusion.

\begin{table}[!t]
\centering
\scalebox{0.9}{
\begin{tabular}{ccc|cc}
\toprule
K & N  & NVPF & mAP & NDS \\ \midrule
1 & 50 & 2.6K & 65.86   & 68.24   \\
3 & 50 & 8.0K & 66.49   & 68.88   \\
6 & 50 & 16.0K & \textbf{66.93}   & \textbf{68.93}   \\ 
10 & 50 & 26.6K & 66.61   & 68.85 \\ 
1 & 200 & 10.6K & 65.59   & 68.00   \\\bottomrule
\end{tabular}}
\caption{Effects of the number of depths per seed (K) and number of seeds per instance (N) in multi-depth unprojection. NVPF represents the number of virtual points per frame.}
\label{table:NDPS}
\end{table}

\begin{figure}[!h]
\centering
\includegraphics[width=\columnwidth]{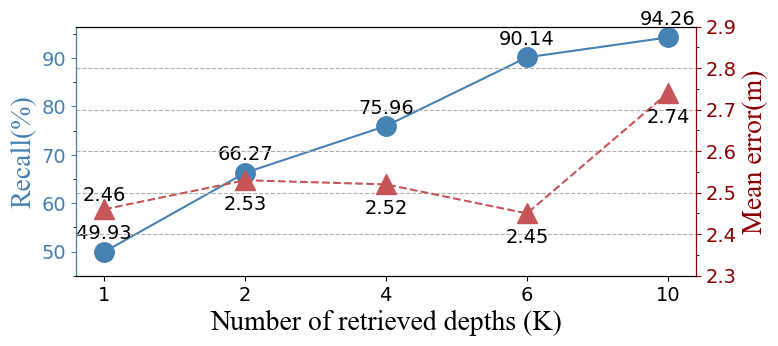}
\caption{Mean error and recall of the virtual points computed with different numbers (K) of retrieved depths per seed.}
\label{fig:Err_Recall}
\end{figure}

\subsubsection{Discussion on MDU}
\label{mdu_discussion}
Our proposed MDU is a simple yet effective strategy to lift 2D seeds from image to 3D space, and we would like to further discuss one critical problem: will the multiple depths for a seed introduce geometric noise and diminish the benefits?
We conducted a quantitative experiment on the nuScenes dataset to investigate this problem. Recall that original point cloud, their projection on image plane (i.e., reference points) and seeds are denoted as $\mathcal{P}$, $\mathcal{R}$ and $\mathcal{S}$, respectively.
We fist preserve a subset $\mathcal{R}^{e}=\{(u^e_i,v^e_i,d^e_i)\}_{i=1}^{N^e}$ of the reference points $\mathcal{R}$, and use all $\{(u^e_i,v^e_i)\}$ as seeds to generate virtual points $\hat{\mathcal{P}}^{e}$ by retrieving real depths from the remaining reference points $\mathcal{R}-\mathcal{R}^{e}$ as done in MDU.
Since the real 3D coordinates $\mathcal{P}^{e}$ of the preserved reference points $\mathcal{R}^{e}$ are available,
we then measure two important metrics for the generated virtual points $\hat{\mathcal{P}}^{e}$: (1) The mean distance error in 3D space between $\hat{\mathcal{P}}^{e}$ and $\mathcal{P}^{e}$, which reflects the overall accuracy of the virtual points' coordinates. 
(2) The recall rate that is the ratio of points in $\mathcal{P}^{e}$ which fall within a small neighborhood\footnote{The neighborhood of a point is defined as a small ball with the radius of the smallest voxel's (0.075m, 0.075m, 0.2m) diagonal length, which is 0.23m. Points of such close proximity are equally treated in the following process due to the voxelization operation.} of at least one point in $\hat{\mathcal{P}}^{e}$, which represents MDU's ability to capture unseen real 3D points.
We also vary the number of depths per seed (K) to show its effect.
As shown in Fig.\ref{fig:Err_Recall}, when the number of K increases from 1 to 6, the recall rate of real points improves dramatically,
meanwhile, the mean distance error does not increase significantly. But a too large value of K (e.g., 10) can inevitably introduce noise.
This proves that with a properly chosen number of depths, MDU does not lead to increased noise in depth estimation and can generate virtual points with reliable depths while also covering more real points than a single-depth strategy. 

To further inspect the effects of K on the final detection performance, we keep other settings unchanged and vary K to compare performances. As Table~\ref{table:NDPS} shows, increasing K from 1 to 6 can gradually improve these models' performances. As we previously analyzed in Fig.\ref{fig:Err_Recall}, such improvements can be mainly attributed to the increased recall and tolerable distance error of the virtual points. However, when 
K increases to 10, the model's performance drops because the extra noise diminishes the benefits of correctly generated virtual points.  
Moreover, we also study whether increasing the number of seeds per instance N can achieve similar improvements, and the results when K is set to 1 and N is set to 200 are shown in the last row of Table~\ref{table:NDPS}. 
It can be seen that simply increasing seeds cannot bring improvements as our multi-depth strategy, but can even hurt performance due to more noise from single-depth unprojection.

\section{Conclusion}
In this paper, we present MSMDFusion, a novel LiDAR-camera fusion framework for 3D object detection which performs multi-modal interaction across multiple scales. We adopt a Multi-Depth Unprojection (MDU) strategy to obtain reliable virtual points with depth-aware semantics from images, and apply Gated Modality-Aware Convolution (GMA-Conv) in each scale to encourage fine-grained controllable multi-modal LiDAR-camera fusion. Multi-granularity information is further combined across scales to form comprehensive features for the prediction head. Extensive experiments demonstrate the effectiveness of these components and our method finally achieves state-of-the-art performances on the nuScenes dataset.

{\small
\bibliographystyle{ieee_fullname}
\bibliography{egbib}
}

\end{document}